\documentclass[11pt,a4paper]{article}
\usepackage[hyperref]{acl2021}
\usepackage{times}
\usepackage{url}
\usepackage{algorithm}
\usepackage{multicol}
\usepackage{algorithmicx}
\usepackage{mathrsfs}
\usepackage{latexsym}
\usepackage{graphicx}
\usepackage{float}
\usepackage{amssymb}
\usepackage{amsmath, bm}
\usepackage{array}

\usepackage{algpseudocode}

\usepackage{microtype}

\usepackage{microtype}

\aclfinalcopy 

\title{Improving Event Causality Identification via Self-Supervised Representation Learning on  External Causal Statement}

\author{Xinyu Zuo$^{1,2}$, Pengfei Cao$^{1,2}$, Yubo Chen$^{1,2}$, Kang Liu$^{1,2}$, Jun Zhao$^{1,2}$, \\
\textbf{Weihua Peng}$^{3}$ \and \textbf{Yuguang Chen}$^{3}$
 \\
	$^1$National Laboratory of Pattern Recognition, Institute of Automation,
 CAS, Beijing, China\\
 $^2$School of Artificial Intelligence, University of Chinese Academy of Sciences, Beijing, China \\
 $^3$Beijing Baidu Netcom Science Technology Co., Ltd \\
	{\tt \{xinyu.zuo,pengfei.cao,yubo.chen,kliu,jzhao\}@nlpr.ia.ac.cn} \\
	{\tt \{pengweihua,chenyuguang\}@baidu.com}
	} 

\date{}

\begin{document}
\maketitle
\begin{abstract}
Current models for event causality identification (ECI) mainly adopt a supervised framework, which heavily rely on labeled data for training. Unfortunately, the scale of current annotated datasets is relatively limited, which cannot provide sufficient support for models to capture useful indicators from causal statements, especially for handing those new, unseen cases. To alleviate this problem, we propose a novel approach, shortly named CauSeRL, which leverages external causal statements for event causality identification. First of all, we design a self-supervised framework to learn context-specific causal patterns from external causal statements. Then, we adopt a contrastive transfer strategy to incorporate the learned context-specific causal patterns into the target ECI model. Experimental results show that our method significantly outperforms previous methods on EventStoryLine and Causal-TimeBank (+2.0 and +3.4 points on F1 value respectively).

\end{abstract}

\section{Introduction}
\begin{figure}[t]
	\centering
	\includegraphics*[clip=true,width=0.45\textwidth,height=0.14\textheight]{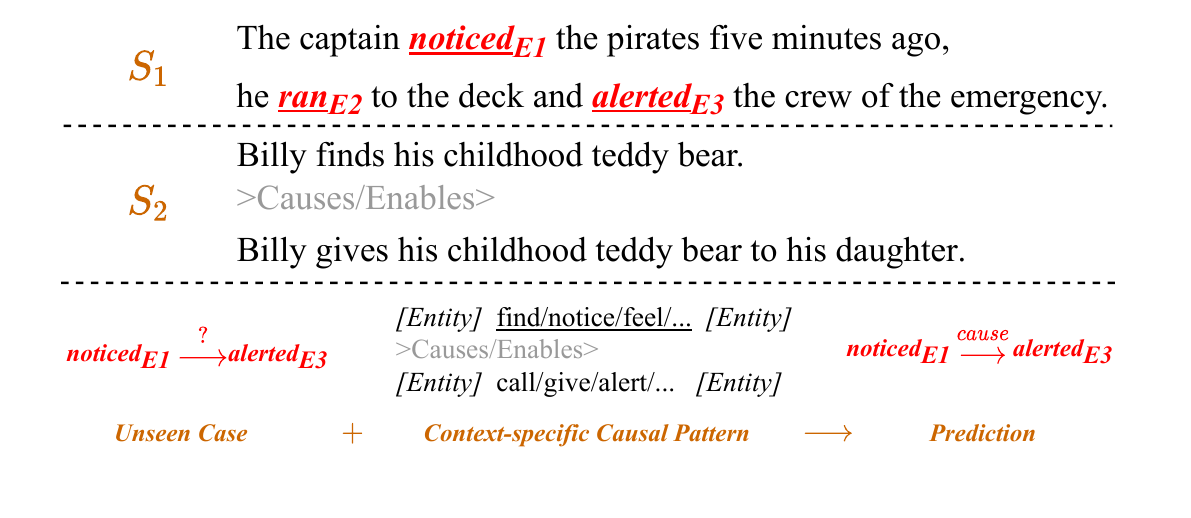}
	\caption{$S_1$ is a labeled data that contains unseen causal events and their statement when training; $S_2$ is an external causal statement; The bottom illustrates the context-specific causal pattern in $S_2$ could help identify the causality of unseen events in $S_1$.} \label{fig1}
\end{figure}

Event causality identification (ECI) aims to identify causal relations between events in texts, which can provide crucial clues for deep textual understanding \cite{girju2003automatic,oh2013question,oh2017multi}. For example in Figure \ref{fig1}, an ECI system should identify two causal relations in $S_1$ with mentioned events: \textbf{noticed$_{E1}$} $ \stackrel{cause}{\longrightarrow}$ \textbf{alerted$_{E3}$} and \textbf{alerted$_{E3}$} $\stackrel{cause}{\longrightarrow}$ \textbf{ran$_{E2}$}. 

To date, most existing methods regard this task as a classification problem and usually train ECI models on annotated data \cite{hashimoto2014toward,riaz2014recognizing,mirza2016catena,hu2017inferring,gao-etal-2019-modeling}. However, the scale of current annotated datasets are relatively limited, where the so far largest dataset EventStoryLine \cite{caselli2017event} only contains 258 documents, 4316 sentences, and 1770 causal event pairs. As a result, on the limited annotated examples, existing ECI models could not easily capture useful indicators from causal statements, especially for handing those new, unseen cases.

To address this problem, \citet{ijcai2020-499} employed external event-related knowledge bases (KBs) to enhance the causality inference, where those KBs store inherent causal relations between some given events. For those unseen events and unlabeled causalities in KBs, \citet{ijcai2020-499} proposed a mention-mask based reasoner to enhance the causal statement representation. However, such mention-mask based reasoner is still trained on the human-annotated examples solely. It will still suffer from data limitations and have no capacity to handling unseen contexts. Moreover, \citet{zuo-etal-2020-knowdis} improved the performance of ECI with the distantly supervised labeled training data. However, their models are still limited to the unsatisfied qualities of the automatically generated data.

To address the insufficient annotated example problem, we employ a large number of external causal statements \cite{DBLP:journals/corr/abs-1811-00146,mostafazadeh-etal-2020-glucose} that can support adequate evidence of context-specific causal patterns \cite{ijcai2020-499} for understanding event causalities. For example in Figure \ref{fig1}, the context-specific causal pattern support by an external causal statement $S_2$ is helpful for identifying the causality of event \emph{noticed$_{E1}$} and event \emph{alerted$_{E3}$} in $S_1$, which is unseen when only training with labeled data. However, different from annotated examples for the ECI task, there are no event annotations in the external causal statements. As a result, it is difficult for the models to learn context-specific causal patterns from them to identify event causalities. To resolve this issue, inspired by \citet{DBLP:conf/nips/GrillSATRBDPGAP20}, we design a self-supervised representation learning framework to learn enhanced causal representations from external causal statements. Specifically, we iteratively sample two external causal statements, then take each of them as a target to learn the commonalities among them. Intuitively, we believe that the learned commonalities between different causal statements through self-supervision reflect such context-specific causal patterns which are helpful for identifying event causalities in the unseen cases. 


Moreover, to incorporate the learned context-specific causal patterns from external causal statements into the target ECI model, we employ a contrastive transfer strategy. In specific, we regard the self-supervised representation learning module as a teacher model that \emph{masters} abundant external causal statements, and the target ECI model as a student model. Methodologically, we make the representation of the causal events encoded by the student model should be close to the causal representation grasped by the teacher model, and keep the representation of the non-causal events away from it. In this way, the mutual information between the teacher and student models could be maximized \cite{Tian2020Contrastive}. Then the learned context-specific causal patterns could be naturally transferred into the ECI model and the generalization could be improved.  

In experiments, we evaluate our model on two benchmarks. The experimental results show that our model achieves SOTA performance. Then, concrete proofs show that the effectiveness of our self-supervised contrast-based framework for context-specific causal patterns learning and transfer.

In summary, the contributions are as follows:
\begin{itemize}
	\item We propose a novel approach, shortly named \textbf{CauSeRL}, which could leverage external causal statements to identify the causalities between events.
	
	\item First of all, we design a self-supervised framework to learn context-specific causal patterns from external causal statements. Then, we adopt a contrastive transfer strategy to incorporate the learned context-specific causal patterns into target ECI model for identification.
	
	\item Experimental results on two benchmarks show that our model achieves the best performance.
\end{itemize}

\section{Related Work}
\begin{figure*}[t]\
	\centering
	\includegraphics*[clip=true,width=0.80\textwidth,height=0.25\textheight]{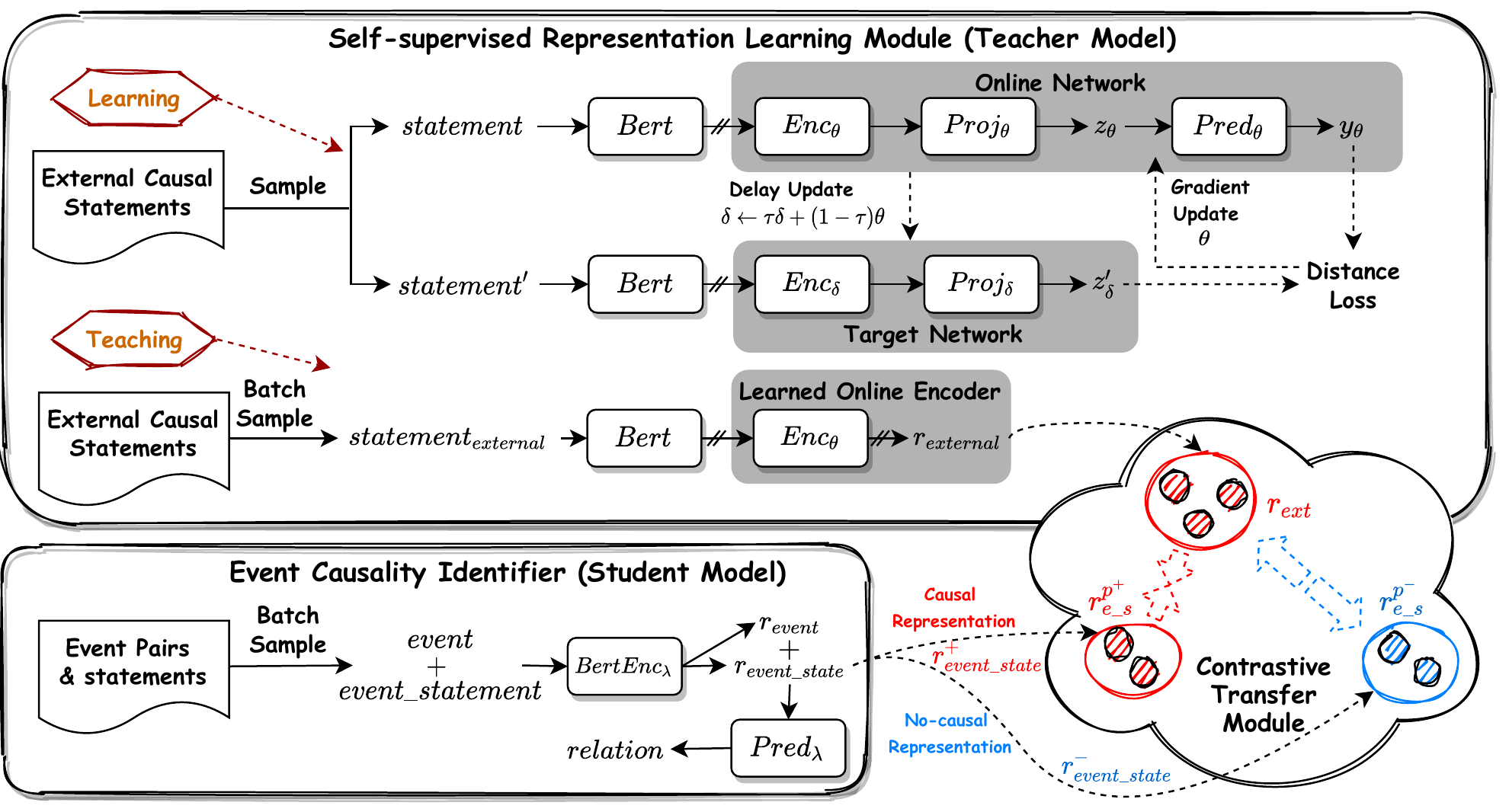}
	\caption{The learning and transfer processes of the proposed CauSeRL for ECI. "$//$" means stop-gradient.} \label{fig2}
\end{figure*}

\paragraph{Event Causality Identification} Up to now, identifying the causality implied in the text has attracted more and more attention \cite{riaz2013toward,riaz2014recognizing,hashimoto2014toward,riaz2014depth,riaz2010another,do2011minimally,hidey-mckeown-2016-identifying,beamer2009using,hu2017inference,hu2017inferring}. 
Recently, some benchmarks on the event causality have been released. \citet{mirza2014annotating}, \citet{mirza2016catena} extracted causal relation of events with a rule-based multi-sieve approach incorporating with event temporal relation. \citet{Mirza2014AnAO} annotated the Causal-TimeBank of event causal relations. \citet{caselli2017event} annotated the EventStoryLine Corpus for event causality identification in 320 short stories based on the temporal and causal relations annotated dataset \cite{mostafazadeh2016caters}. \citet{dunietz-etal-2017-corpus} presented BECauSE 2.0, a new version of the BECauSE \cite{dunietz-etal-2015-annotating} of causal relation and other seven relations. 

Based on the above benchmarks, \citet{gao-etal-2019-modeling} modeled document-level structures to identify the causalities of events. \citet{ijcai2020-499} identified event causalities with the mention masking generalization and external KBs. \citet{zuo-etal-2020-knowdis} improved the performance of ECI with the distantly automatically labeled training data. However, these methods only rely on a small scale of labeled data. In this paper, we introduce external causal statements to help identify event causalities.

\paragraph{Self-Supervised Representation Learning} Self-supervised representation learning cares about producing good features generally helpful for many tasks \cite{weng2019selfsup}. \citet{wu2018unsupervised} proposed MemoryBank, which stores representations of all the data and samples a random set of keys as negative examples. \citet{he2020momentum} provided a framework, MoCo, of unsupervised learning visual representation as a dynamic dictionary look-up. \citet{chen2020simple} proposed the SimCLR which learns representations for visual inputs by maximizing agreement between differently augmented views of the same sample via a contrastive loss. \citet{DBLP:conf/nips/GrillSATRBDPGAP20} claimed a novel representation learning framework relies on two neural networks, BYOL, without using negative samples. CURL \cite{srinivas2020curl} applies the above ideas in reinforcement learning. Inspired by them, we design a self-supervised framework to learn context-specific causal patterns from external causal statements and adopt a contrastive transfer strategy to incorporate them into target ECI model.

\section{Methodology}
As shown in Figure \ref{fig2}, the whole pipeline process of CauSeRL is divided into two major stages. 

\begin{itemize}
    \item \textbf{Self-supervised causal representation learning}  (\textbf{SelfRL}, Sec. \ref{sec:CRL}). In this stage, we design a \emph{self-supervised representation learning module} to learn enhanced causal representations by iteratively sampling two external causal statements, taking each of them as a target to learn their commonalities which reflect context-specific causal patterns.
    
    \item \textbf{Contrastive representation transfer} (\textbf{ConRT}, Sec. \ref{sec:CGT}). In this stage, we employ a \emph{contrastive transfer module} to transfer the learned context-specific causal patterns into the ECI target model, the \emph{event causality identifier}, via incorporating the enhanced causal representations from SelfRL.
\end{itemize}


\subsection{Self-Supervised Causal Representation Learning (SelfRL)}
\label{sec:CRL}
SelfRL aims to train a module that masters context-specific causal patterns from external causal statements by learning their enhanced causal representation with a self-supervised framework.

\paragraph{Self-Supervised Representation Learning Module}
\label{sec:TM}
We design a self-supervised module to capture the context-specific causal patterns from external causal statements via learning their enhanced causal representation. However, there are no ECI-specific event annotations in the external causal statements, which makes them unable to be directly used as training data to train the ECI model. To handle this problem, inspired by \citet{DBLP:conf/nips/GrillSATRBDPGAP20}, we iteratively sample two external causal statements, take each of them as a target to learn their commonalities, that is, the causal representations, which reflect context-specific causal patterns.

In specific, as shown in Figure \ref{fig2}, we configure two networks for SelfRL, an online network, and a target network. The target network provides regression targets to train the online network which makes it learn the commonalities among two input causal statements, that is, the causal representations reflecting different context-specific causal patterns. Structurally, the online network is defined as a set of weights $\theta$ which is comprised of three submodules: an \emph{encoder} $Enc_\theta$, a \emph{projector} $Proj_\theta$ and a \emph{predictor} $Pred_\theta$. And the target network has the same architecture as the online network, but no predictor and uses a different set of weights $\delta$. 

In specific, we iteratively sample two external causal statements, initially encode them by BERT \cite{devlin-etal-2019-bert}, and input them into two networks respectively. After encoding and projection, the online network and target network respectively output a projection $\bm{z}_{\theta}$ and $\bm{z}_{\delta}^{\prime}$. Then the online network outputs a prediction $\bm{y}_{\theta}$, and takes the following mean square error between $\ell_{2}$-normalized $\bm{\bar{y}}_{\theta}$ and $\bm{\bar{z}}_\delta^{\prime}$ as the training objective to learn the commonalities of two causal statements, that are regarded as the context-specific causal patterns.
\begin{align} \footnotesize
\mathcal{L}_{\theta, \delta} 
    \triangleq \left\|\bm{\bar{y}}_{\theta} -\bm{\bar{z}}_\delta^{\prime}\right\|_{2}^{2}
     = & 2 - 2 \cdot \frac{\left\langle \bm{y}_{\theta}, \bm{z}_{\delta}^{\prime}\right\rangle}{\left\|\bm{y}_{\theta}\right\|_{2} \cdot\left\|\bm{z}_{\delta}^{\prime}\right\|_{2}}, \\
    \bm{\bar{y}}_{\theta} \triangleq \bm{y}_{\theta} /\left\|\bm{y}_{\theta}\right\|_{2} &, \bm{\bar{z}}_{\delta}^{\prime} \triangleq \bm{z}_{\delta}^{\prime} /\left\|\bm{z}_{\delta}^{\prime}\right\|_{2}.
\end{align}
To reduce the bias, we symmetrize the $\mathcal{L}_{\theta, \delta}$ by swapping the input causal statements of the online and target networks to compute $\widetilde{\mathcal{L}}_{\theta, \delta}$.

\paragraph{Learning of SelfRL}
For the learning of SelfRL, at each step, as shown in Algorithm \ref{alg1}, we minimize the $\mathcal{L}_{\theta, \delta}^{tea}$ to stochastic gradient update the online network respect to the parameters $\theta$ only. For the target network, the parameters $\delta$ are an exponential moving average of the parameters $\theta$ of the online network \cite{Lillicrap2016ContinuousCW}:
\begin{align} \footnotesize
    \mathcal{L}_{\theta, \delta}^{tea} & = \mathcal{L}_{\theta, \delta} + \widetilde{\mathcal{L}}_{\theta, \delta}, \\
    \theta & \leftarrow \eta_{tea} \nabla_{\theta} \mathcal{L}_{\theta, \delta}^{tea}, \\
    \delta & \leftarrow \tau \delta + (1 - \tau) \theta,
\end{align}
where, $\eta_{tea}$ is the learning rate of the online network, and $\tau \in [0,1]$ is the decay rate that determines the degree of the movement of $\theta$ to $\delta$. As shown in Figure \ref{fig2}, when learning, BERT is only used to provide an initial representation for the input statements, and its parameters are not updated.

According to the theoretical analysis by \citet{DBLP:conf/nips/GrillSATRBDPGAP20}, the addition of a predictor on the online network and the usage of a slow-moving average of the online parameters as the target network encourage SelfRL to encode a more informative causal representation of commonalities within the online projection and avoids collapsed solutions\footnote{In this paper, collapse solution means that the model encodes all input statements as the same representation. The slow-moved target network keeps the predictor of the online network always near-optimal, thus avoiding the collapse.}.

\begin{algorithm}[t] \footnotesize
	\caption{Two stages training of CauSeRL.}
	\begin{algorithmic}[1]
		\Require External causal statements $\mathcal{C}$ for teacher model $\quad$ and event pairs with statements $\mathcal{P}$ for student model.
		\Ensure 
		
		\Function {Causal Representation Learning}{} 
		\For{each batch $\mathcal{C}_{bat} \in \mathcal{C}$} \Comment{\emph{Learning of SelfRL}}
		\For{any two causal statements $\in \mathcal{C}_{bat}$} 
		\State One for online another for target;
		\State Get $y_{\theta}$ from $Pred_{\theta}$ in online network;
		\State Get $z_{\delta}^{\prime}$ from $Proj_{\delta}$ in target network;
		\State Swap two statements into two networks;
		\State Get symmetrical $y_{\theta}$ and $z_{\delta}^{\prime}$;
		\State Compute $\mathcal{L}_{\theta, \delta}$ and $\widetilde{\mathcal{L}}_{\theta, \delta}$;
		\EndFor
		\State Compute batch $\mathcal{L}_{\theta, \delta}^{tea}$ in equation (3);
		\State Stochastic gradient update $\theta$ in equation (4);
		\State Slow-moving update $\delta$ in equation (5);
		\EndFor
		\EndFunction  
		\\
		\Function {Contrastive Representation Transfer}{}
		\For{each batch $\mathcal{P}_{bat} \in \mathcal{P}$} \Comment{\emph{Learning of identifier}}
		\For{any event pair with statement $\in \mathcal{P}_{bat}$} 
		\State Get $\bm{r}_{event}$ and $\bm{r}_{event\_state}$ from $BertEnc_{\lambda}$;
		\State Predict the causality of two events in one pair;
		\EndFor
		\State Compute batch $\mathcal{L}_{\lambda}^{stu}$ in equation (6);
		\State Sample $\mathcal{C}_{bat} \in \mathcal{C}$;
		\State Get $\bm{r}_{external}$ of $c \in \mathcal{C}_{bat}$ from learned $Enc_{\theta}$;
		\State Get $\bm{r}_{event\_state}^{+}$, $\bm{r}_{event\_state}^{-}$ from $\bm{r}_{event\_state}$;
		\State Get mapped $\bm{r}_{e\_s}^{p^+}$, $\bm{r}_{e\_s}^{p^-}$ and $\bm{r}_{ext}$;
		\State Compute $\mathcal{L}_{\lambda}$ = $\mathcal{L}_{\lambda}^{stu}$ + $\mathcal{L}_{\lambda}^{con}$ in equation (8);
		\State Stochastic gradient update $\lambda$ in equation (9);
		\EndFor
		\EndFunction 
	\end{algorithmic}
	\label{alg1}
\end{algorithm}

\begin{table*}[t] \footnotesize
\centering
\begin{tabular}{|m{1.2cm}|m{7.2cm}|m{5.7cm}|}
\hline
\multicolumn{1}{|c|}{\textbf{Resource}}    & \multicolumn{1}{|c|}{\textbf{Original Causal Statement Form}}                                                                         & \multicolumn{1}{|c|}{\textbf{Converted Causal Statement Form}}                                                    \\ \hline
\multicolumn{1}{|c|}{\textbf{GLU-SPE}} & Billy finds his childhood teddy bear $>Cause/Enable>$ Billy gives his childhood teddy bear to his daughter & Billy finds his childhood teddy bear, billy gives his childhood teddy bear to his daughter. \\ \hline
\multicolumn{1}{|c|}{\textbf{GLU-GEN}} & Someone\_A finds Something\_A $>Cause/Enable>$ Someone\_A gives Something\_A to Someone\_B                         & Someone\_A finds Something\_A, Someone\_A gives Something\_A to Someone\_B.                       \\ \hline
\multicolumn{1}{|c|}{\textbf{ATOMIC}}      & PersonX follows PersonY into room $>oWant>$ to know why PersonX is following them                             & PersonX follows PersonY into room, to know why PersonX is following them.                   \\ \hline
\multicolumn{1}{|c|}{\textbf{DISTANT}}     & Fisk was shot to death by his mistress's new lover and Fisk's ex-business partner.                          & Fisk was shot to death by his mistress's new lover and Fisk's ex-business partner.           \\ \hline
\end{tabular}
\caption{The original and converted form (the input form of SelfRL) of different causal statements from three resources. GLU-SPE and GLU-GEN denote the specific and general statements from GLUCOSE respectively.}
\label{tab1}
\end{table*}

\subsection{Contrastive Representation Transfer (ConRT)}
\label{sec:CGT}
ConRT aims to incorporate the context-specific causal patterns learned in SelfRL from external causal statements into the identifier. As aforementioned, the goal of SelfRL is learning the commonalities among different external causal statements, which does not make the representation learning module have the ability to distinguish the causal and non-causal statements directly. Therefore, we employ a contrastive transfer module to teach the learned context-specific causal patterns to the event causality identifier for training.

\paragraph{Event Causality Identifier}
\label{sec:SM}
Event causality identification is formulated as a sentence level binary classification problem. Specifically, we design a classifier based on BERT \cite{devlin-etal-2019-bert} to build our identifier. The input is an event pair and its statement. As shown in Figure \ref{fig2}, we take representation of events $\bm{r}_{event}$ and their contextual statement $\bm{r}_{event\_state}$ encoded by $BertEnc_{\lambda}$ as the input of top MLP predictor. Finally, the output is a binary vector to indicate the causal relation of the input two events expressed by their statement. The parameters of the identifier are defined as $\lambda$ and the optimization function is the following classification cross-entropy function:
\begin{align} \footnotesize
    \mathcal{L}_{\lambda}^{stu} = \textsc{CrossE}(\textsc{MLP}([\bm{r}_{event};\bm{r}_{event\_state}])).
\end{align} 

\paragraph{Contrastive Transfer Module}
\label{sec:CS}
As aforementioned, inspired by \citet{Tian2020Contrastive}, we employ a contrastive transfer strategy to transfer the \emph{"knowledge"} mastered by the teacher (self-supervised representation learning module), that is the context-specific causal patterns, to the student (event causality identifier), which helps the latter to identify the event causalities. The key idea of contrastive transfer is intuitional: maximize the mutual information between the teacher and the student \cite{Tian2020Contrastive}. Methodologically, we make the representation of the statements of causal events encoded by the student model should be close to the causal representation grasped by the teacher model. By contrast, we keep the representation of the statements of non-causal events away from it.

As shown in Figure \ref{fig2}, at each training step of identifier, we sample a batch of external causal statements into the learned $Enc_{\theta}$ of the online network to obtain their causal representation $\bm{r}_{ext}$ for teaching. At the same time, we also sample a batch of event pairs with their statements into the $BertEnc_{\lambda}$ of identifier to obtain the statement representation $\bm{r}_{event\_state}$ of each event pair. Among one batch, $\bm{r}_{event\_state}$ consists of the $\bm{r}_{event\_state}^+$ of causal event pairs and the $\bm{r}_{event\_state}^-$ of non-causal event pairs. After mapping $\bm{r}_{external}$, $\bm{r}_{event\_state}^+$ and $\bm{r}_{event\_state}^-$ into a same space, we obtain $\bm{r}_{ext}$, $\bm{r}^{p^+}_{e\_s}$ and $\bm{r}^{p^-}_{e\_s}$ respectively. After that, we make $\bm{r}^{p^+}_{e\_s}$ be close to $\bm{r}_{ext}$ in the contrastive loss function: 

\begin{equation} \footnotesize
\mathcal{L}^{con}_{\lambda} = \frac{1}{|\mathcal{P}^+|} \sum_{p^+ \in \mathcal{P}^+} \log \frac{e^{(\mathcal{D}(\bm{r}^{p^+}_{e\_s}, \bm{r}_{ext}) / T)}}{\sum_{p \in P} e^{(\mathcal{D}(\bm{r}^{p}_{e\_s}, \bm{r}_{ext}) / T)}},
\end{equation}
where, $P^+$ and $P$ are the causal event pairs and all event pairs in one batch respectively, $T$ is a temperature that adjusts the concentration level, 
and $\mathcal{D}$ is the $\ell_{2}$-distance function to measure the distance of two representation.

\paragraph{Learning of Event Causality Identifier}
For the training of event causality identifier, we add contrastive loss to the basic classification loss, which could guide the identifier to learn context-specific causal patterns implied in the enhanced causal representation from SelfRL. As shown in Algorithm \ref{alg1}, we minimize the $L_{\lambda}$ and stochastic gradient update the $\lambda$ as following:
\begin{align} \footnotesize
    \mathcal{L}_{\lambda} & = \mathcal{L}_{\lambda}^{stu} + \mathcal{L}^{con}_{\lambda}, \\
    \lambda & \leftarrow \eta_{stu} \nabla_{\lambda} \mathcal{L}_{\lambda},
\end{align}
where, $\eta_{stu}$ is the learning rate of the identifier. For evaluation, we predict the causality of input event pair without the contrastive transfer module. Additionally, the $T$ in $\mathcal{L}^{con}_{\lambda}$ indirectly plays a role in adjusting the influence weight of $\mathcal{L}_{\lambda}^{stu}$ and $\mathcal{L}^{con}_{\lambda}$. In specific, for teaching, we take the learned $Enc_{\theta}$ of the online network as the encoder, freeze its parameters, to provide the enhanced causal representation of the external causal statements for contrastive representation transfer.

\section{Experiments}
\subsection{Experimental Setup}
\paragraph{Dataset and Evaluation Metrics for ECI}
Our experiments are conducted on two main benchmarks, including: \textbf{EventStoryLine} v0.9 (ESC) \cite{caselli2017event} described above; and (2) \textbf{Causal-TimeBank} (CTB) \cite{Mirza2014AnAO} which contains 184 documents, 6813 events, and 318  causal event pairs. Same as previous methods, we use the last two topics of ESC as the development set for two datasets. For evaluation, we adopt Precision (P), Recall (R), and F1-score (F1) as evaluation metrics. We conduct 5-fold and 10-fold cross-validation on ESC and CTB respectively, same as previous methods. All the results are the average of three independent experiments. 

\paragraph{Data Preparation for Self-Supervised Causal Representation Learning} 
We take four types of external causal statements from three resources. Table \ref{tab1} illustrates the original form and the converted input form of SelfRL (Sec. \ref{sec:CRL}) of the causal statements from three different resources.
\begin{itemize}
    \item \textbf{GLUCOSE} \cite{mostafazadeh-etal-2020-glucose}: a large-scale dataset of implicit commonsense knowledge, encoded as causal explanatory mini-theories inspired by cognitive psychology. Each GLUCOSE explanation is stated both as a specific statement (grounded in a given context, \textbf{GLU-SPE} in Table \ref{tab1}) and a corresponding general rule (applicable to other contexts, \textbf{GLU-GEN} in Table \ref{tab1}).
    
    \item \textbf{ATOMIC} \cite{DBLP:journals/corr/abs-1811-00146}: an atlas of machine commonsense, as a step toward addressing the rich spectrum of inferential knowledge that is crucial for commonsense reasoning.
    
    \item \textbf{DISTANT} \cite{zuo-etal-2020-knowdis}: the automatically labeled training data for ECI via distant supervision that expresses the causal semantics between events.
\end{itemize}

	
\paragraph{Parameters Settings} 
In implementations, all the BERT modules are implemented on BERT-Base architecture\footnote{\url{https://github.com/google-research/bert}}, which has 12-layers, 768-hiddens, and 12-heads. We employ the one-layer BiLSTM \cite{Bi-LSTM} as $Enc_{\theta}$ and $Enc_{\delta}$. For parameters, we set the learning rate of SelfRL ($\eta_{tea}$) and identifier ($\eta_{stu}$) as 1e-5 and 2e-5 respectively. The size of the space in the contrastive transfer module and the hidden layer of BiLSTM are both set as 50. And we respectively set the decay rate $\tau$ of moving average in SelfRL and the temperature of the contrastive loss $\mathcal{L}^{con}_{\lambda}$ are 0.996 and 0.1 tuned on the development set. Moreover, we also tune the batch size of SelfRL and identifier as 48 and 16 respectively on the development set. And we apply the early stop and AdamW gradient strategy to optimize all models. We also adopt a negative sampling rate of 0.6 for the training of identifier, owing to the sparseness of positive examples in the ECI datasets.

\paragraph{Compared Methods} Same as previous methods. For ESC, we prefer 1) \textbf{S-Path} \cite{cheng2017classifying}, a dependency path based sequential method that models the context between events to identify causality; 2) \textbf{S-Fea} \cite{choubey2017sequential}, a sequence model explores complex human designed features for ECI; 3) \textbf{LR+} and \textbf{ILP} \cite{gao-etal-2019-modeling}, document-level models adopt document structures for ECI. 

For CTB, we prefer 1) \textbf{Rule-B}, a rule-based system; 2) \textbf{Data-D}, a data driven machine learning based system; 3) \textbf{VerR-C}, a verb rule based model with data filtering and causal signals enhancement. These models are designed by Mirza and Tonelli \shortcite{Mirza2014AnAO,DBLP:journals/corr/Mirza16} for ECI. For both two datasets, 1) we build a baseline \textbf{BERT} (our basic proposed event causality identifier); 2) We prefer \textbf{MasG} \cite{ijcai2020-499}, a BERT-Large based SOTA model with mention masking generalization; 3) \textbf{KnowDis} \cite{zuo-etal-2020-knowdis} improved the performance of ECI with the distantly labeled training data.

\begin{table}[t] \footnotesize
	\centering
	\scalebox{0.99}{
	\begin{tabular}{lccc}
		\textbf{Methods}    & \textbf{P} & \textbf{R} & \textbf{F1}  \\ \hline
		\multicolumn{4}{c}{\textbf{EventStoryLine}} \\ \hline
		S-Path \cite{cheng2017classifying}         & 34.0       & 41.5       & 37.4  \\ 
		S-Fea \cite{choubey2017sequential}         & 32.7       & 44.9       & 37.8 \\ 
		LR+ \cite{gao-etal-2019-modeling}        & 37.0       & 45.2       & 40.7  \\ 
		ILP \cite{gao-etal-2019-modeling}        & 37.4       & 55.8       & 44.7   \\ 
		BERT      & 36.0       & 56.8       & 44.1       \\ 
		KnowDis \cite{zuo-etal-2020-knowdis} &  39.7 &  66.5  &  49.7  \\ 
		MasG \cite{ijcai2020-499} &  41.9 &  62.5  &  50.1  \\ \hline
		KnowDis+CauSeRL (\textbf{Ours}) &  40.1 &  68.9  &  50.7* \\
		MasG+CauSeRL (\textbf{Ours}) &  40.8 &  68.0  &  51.0* \\ \hline
		CauSeRL$_{DISTANT}$ (\textbf{Ours})    & 39.9       & 67.3       & 50.1*  \\
		 CauSeRL$_{ATOMIC}$ (\textbf{Ours})    & 41.0       & 68.1       & 51.2*               \\
		  CauSeRL$_{GLU\text{-}GEN}$ (\textbf{Ours})   & 41.4       & 67.8       & 51.4*      \\
		CauSeRL$_{GLU\text{-}SPE}$ (\textbf{Ours})  & \textbf{41.9}       & \textbf{69.0}       & \textbf{52.1}*               \\ \hline
		\multicolumn{4}{c}{\textbf{Causal-TimeBank}}    \\ \hline
		Rule-B \cite{Mirza2014AnAO}  &  36.8  &  12.3  & 18.4  \\ 
		Data-D \cite{Mirza2014AnAO}  &  67.3  &  22.6  & 33.9  \\
		VerR-C \cite{DBLP:journals/corr/Mirza16}  &  \textbf{69.0}  &  31.5  & 43.2    \\ 
		BERT     & 39.5  & 44.5  &  41.9  \\
		MasG \cite{ijcai2020-499}    & 36.6  &  55.6  &  44.1  \\ 
		KnowDis \cite{zuo-etal-2020-knowdis}    & 42.3  &  60.5  &  49.8  \\ \hline
		MasG+CauSeRL (\textbf{Ours})   & 42.6  & 62.5  &  50.7*  \\ 
		KnowDis+CauSeRL (\textbf{Ours}) &  42.5 &  66.0  &  51.7*  \\  \hline
		CauSeRL$_{DISTANT}$ (\textbf{Ours})    & 41.6       & 63.9       & 50.4*  \\
		 CauSeRL$_{ATOMIC}$ (\textbf{Ours})    & 42.8       & 67.0       & 52.2*              \\
		  CauSeRL$_{GLU\text{-}GEN}$ (\textbf{Ours})  & 43.0       & 66.8       & 52.3*        \\
		CauSeRL$_{GLU\text{-}SPE}$ (\textbf{Ours})  &   43.6    &  \textbf{68.1}   &  \textbf{53.2*}       \\ \hline
	\end{tabular}}
	\caption{Results of event causality identification on two benchmarks. Bold denotes best results; * denotes a significant test at the level of 0.05; 
	} 
	\label{tab2} 
\end{table}
	
To make a fair comparison, we employ CauSeRL to retrain MasG and KnowDis to illustrate the effectiveness of our proposed approach for ECI on other methods. In specific, 1) \textbf{MasG+CauSeRL}: we retrain MasG with $L_{\lambda}^{con}$ based on the CLU-SPE. To be consistent with other BERT-based compared models, we re-construct MasG based on BERT-Base rather than the original BERT-Large of MasG; 2) \textbf{KnowDis+CauSeRL}: we regard the automatically distantly labeled causal sentences generated by KnowDis as causal statements to learn in SelfRL, and transfer to KnowDis. 

\textbf{CauSeRL$_{External\text{-}Statement}$}: To further illustrate the ability of CauSeRL to learn the context-specific causal patterns for the ECI task, we make CauSeRL learn from four types of external causal statements shown in Table \ref{tab1} for identifying the causalities between events. $External\text{-}Statement$ denotes what kind of external causal statements.

\subsection{Our Method vs. State-of-the-art Methods}
Table \ref{tab2} shows the results of ECI on EventStoryLine and Causal-TimeBank. From the results:

1) Our CauSeRL outperforms all baseline methods and achieves the best performance on F1 value, 52.1\% on ESC and 53.2\% on CTB respectively. Specifically, CauSeRL outperforms the no-bert (ILP/VerR-C) and bert (MasG/KnowDis) baseline methods by a margin of 7.4\%/10.0\% and 2.0\%/3.4\% on two benchmarks respectively. It illustrates the context-specific causal patterns from external causal statements are effective for ECI.  

\begin{table}[t] \footnotesize
\centering
\begin{tabular}{lcccc}
\textbf{Methods}    & \textbf{P} & \textbf{R} & \textbf{F} & \bm{$\nabla$} \\ \hline
CauSeRL$_{GLU\text{-}SPE}$  & \textbf{41.9}         & \textbf{69.0}        & \textbf{52.1*}      & -      \\
${Enc_{\theta-init}}$ + ConRT    & 39.1       & 63.6       & 48.4*       & -3.7           \\
${BertEnc_{\lambda-init}}$ + ConRT    & 38.9       & 63.1       & 48.1*       & -4.0           \\ \hline
BERT       & 36.0       & 56.8       & 44.1       & -              \\
BERT+SelfRL$_{finetune}$  & 38.5       & 60.9       & 47.2*       & +3.1           \\ \hline
\end{tabular}
\caption{Ablation results of the self-supervised causal representation learning (SelfRL, Sec. \ref{sec:CRL}) of ECI on EventStoryLine. * denotes a significant test at the level of 0.05; \bm{$\nabla$} means the points lower than CauSeRL or higher than BERT in the upper and lower parts respectively; ${Enc_{\theta-init}}$ + ConRT denotes a varietal CauSeRL that removes SelfRL, directly employs an initial $Enc_{\theta}$ of the online network to encode external causal statements into ConRT and trains it meanwhile; ${BertEnc_{\lambda-init}}$ + ConRT denotes a varietal CauSeRL that removes SelfRL, directly employs a same initial $BertEnc_{\lambda}$ of identifier to encode external causal statements into ConRT  and trains it meanwhile; BERT+SelfRL$_{finetune}$ denotes a varietal CauSeRL that removes ConRT (Sec. \ref{sec:CGT}), and takes the learned $Enc_{\theta}$ of the online network as the initial encoder of identifier on the BERT baseline model.}
\label{tab3}
\end{table}

2) Comparing MasG+CauSeRL with MasG, we note that even with BERT-Base, the performance of MasG+CauSeRL is significantly higher than that of MasG based on BERT-Large. This shows that the context-specific causal patterns learned by CauSeRL from external causal statements can effectively alleviate the limitation of mask generalization only relying on limited labeled causal context.

3) Comparing KnowDis+CauSeRL with Know- Dis, we find that CauSeRL could more efficiently make use of the automatically labeled causal statements, which learns their context-specific causal patterns to further enhance the ability of models to identify the causalities between events.

4) Comparing different external causal statements. a) GLU-SPE brings the most significant improvement because the specific causal statements from GLU-SPE have complete text structures that are more similar to ECI labeled data and make models easier to learn. There, all the ablation experiments are conducted on GLU-SPE. b) The effects of GLU-GEN and ATOMIC are similar because these two types of statements are abstract causal structures. Although they are similar to the context-specific causal patterns, it is relatively difficult to understand directly. c) The improvement brought by DISTANT is relatively small because of the effects of the noise from distantly labeled data.

\begin{table}[t] \footnotesize
\centering
\begin{tabular}{lcccc}
\textbf{Methods}    & \textbf{P} & \textbf{R} & \textbf{F} & \bm{$\nabla$} \\ \hline
CauSeRL$_{GLU\text{-}SPE}$  & \textbf{41.9}         & \textbf{69.0}        & \textbf{52.1*}      & -           \\
${Enc_{\theta-freeze}}$ + SelfRL & 37.8      & 59.9       & 46.4*       & -5.7           \\
${Enc_{\theta-finetune}}$  + SelfRL & 38.5       & 60.9       & 47.2*       & -4.9           \\ \hline
BERT       & 36.0       & 56.8       & 44.1       & -              \\
BERT + ConRT$_{Enc_{\theta}}$    & 39.1       & 63.6       & 48.4*       & +4.3           \\ \hline
\end{tabular}
\caption{Ablation results of the contrastive representation transfer (ConRT, Sec. \ref{sec:CGT}) of ECI on EventStoryLine. * denotes a significant test at the level of 0.05; \bm{$\nabla$} means the points lower than CauSeRL or higher than BERT in the upper and lower parts respectively; ${Enc_{\theta-freeze}}$ + SelfRL denotes a varietal CauSeRL that removes ConRT, and takes the frozen learned $Enc_{\theta}$ of the online network as the encoder of identifier; ${Enc_{\theta-finetune}}$ denotes a varietal CauSeRL that removes ConRT, and takes the learned $Enc_{\theta}$ of the online network as the initial encoder of identifier; BERT + ConRT$_{Enc_{\theta}}$ denotes a varietal CauSeRL that removes SelfRL (Sec. \ref{sec:CRL}), directly employs an initial $Enc_{\theta}$ of the online network to encode external causal statements into ConRT and trains it meanwhile.}
\label{tab4}
\end{table}

5) Comparing CauSeRL with MasG+CauSeRL, we notice that after removing the ConceptNet knowledge enhancement employed by MasG, the external causal statements could be better learned and transferred. This is because MasG directly flattens the event concept knowledge into the statement sequence, which disrupts the statement structure and affects the understanding of the statement.

6) It is worth noting that the improvement on the CTB is higher than that of the ESC, because the amount of labeled data of the former is relatively small, and more need for the help of external causal statements. Moreover, compared with the traditional methods based on features or rules, all BERT-based methods demonstrate high recall value, which is benefited from more training data, knowledge and causal statements.

\subsection{Effect of Self-Supervised Causal Representation Learning}
We analyze the effect of the self-supervised causal representation learning (SelfRL, Sec. \ref{sec:CRL}). As shown in Table \ref{tab3}, from the results, 1) after removing SelfRL, the performance of ECI significantly decreases. This illustrates that the context-specific causal patterns learned by SelfRL are important for the ECI model to understand the causality. 2) Comparing BERT+SelfRL$_{finetune}$ with BERT, the $Enc_{\theta}$ that has learned from external causal statements could improve the performance of ECI to a certain extent. This illustrates that SelfRL could effectively capture the context-specific causal patterns in the statements for identification. 3) Comparing ${Enc_{\theta-init}}$ + ConRT and ${BertEnc_{\lambda-init}}$ + ConRT, after representation learning, the fine-tuned $Enc_{\theta}$ could further improve the performance of ECI. This indirectly shows that the context-specific causal patterns learned in the SelfRL is generalized.


\begin{figure}[t]
	\centering
	\includegraphics*[clip=true,width=0.40\textwidth,height=0.15\textheight]{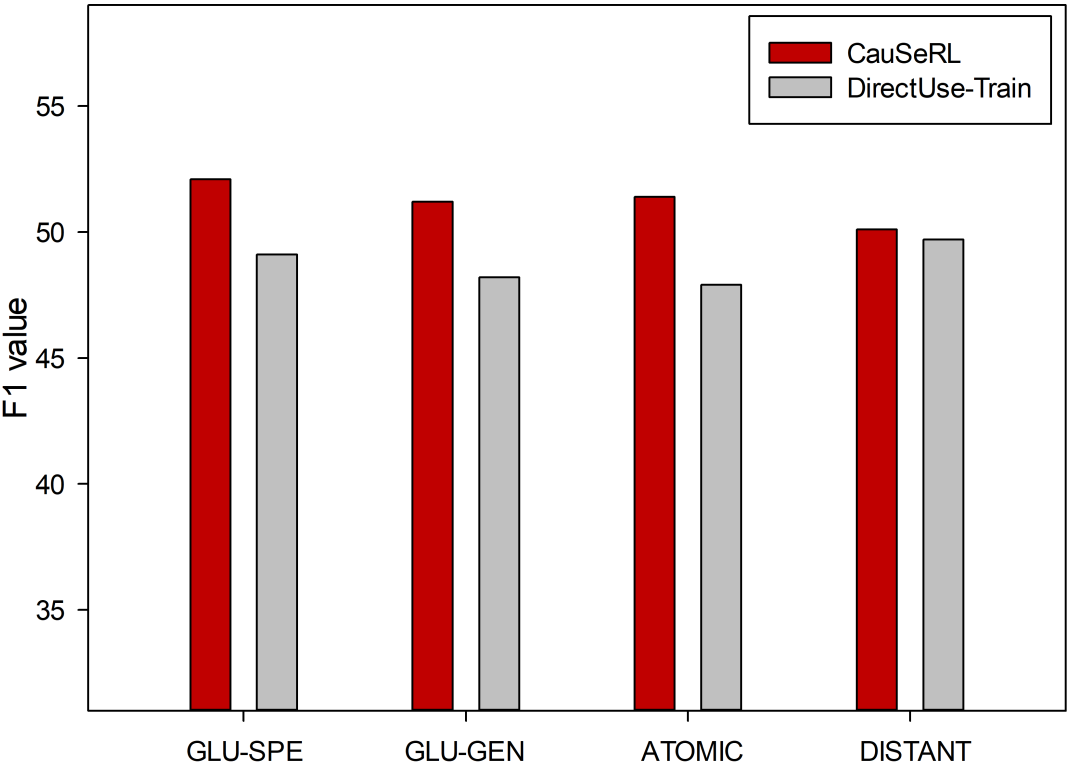}
	\caption{Results of event causality identification on EventStoryLine that directly using external causal statements as the training data of ECI task.} \label{fig3}
\end{figure}

\subsection{Effect of Contrastive Representation Transfer}
We analyze the effect of the contrastive representation transfer (ConRT, Sec. \ref{sec:CGT}). As shown in Table \ref{tab4}, from the results, 1) after removing ConRT, the performance of ECI also significantly decreases. This illustrates that the learned causal representations from external statements are not suitable for direct application to ECI, and needs to be effectively transferred that the ConRT focuses on. 2) Comparing BERT + ConRT$_{Enc_{\theta}}$ with BERT, even if causal representation learning is not carried out in advance, adopting contrast strategy to directly transfer the context-specific causal patterns could also help the inference of event causality to a certain extent. 3) Comparing ${Enc_{\theta-freeze}}$ + SelfRL with ${Enc_{\theta-finetune}}$ + SelfRL, we find that the causal representations encoded by pre-trained BERT and BiLSTM have similar effects. Aforementioned, to avoid collapse solutions (Sec. \ref{sec:CRL}), we choose the BiLSTM as an encoder in SelfRL that could be initialized completely independently.

\subsection{Effect of the Utilization of External Causal Statement}

As shown in Figure \ref{fig3}, we regard external causal statements as positive training data for ECI and directly use them to train the BERT baseline model. In specific, we treat two words that play a predicate role in the syntactic structure of each statement as events. From the results, CauSeRL could more effectively make use of causal statements to help understand the causalities of events. In contrast, directly serving as training data is not effective. 

\subsection{Case Study}
\begin{figure}[t]
	\centering
	\includegraphics*[clip=true,width=0.42\textwidth,height=0.13\textheight]{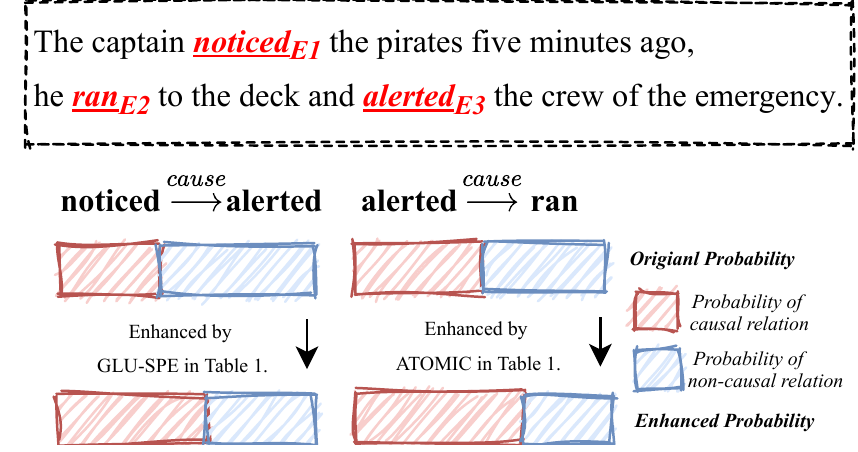}
	\caption{Case study of the probability changes with external causal statements enhancement.} \label{fig4}
\end{figure}

As shown in Figure \ref{fig4}, with limited labeled data, the model could not understand the causal relation between event \emph{noticed} and event \emph{alerted}. Fortunately, with the support of the context-specific causal pattern from GLU-SPE in Table \ref{tab1}, the prediction is modified correctly. Moreover, the original model that only trained with limited labeled data is ambiguous about the causal relation between event \emph{alerted} and event \emph{ran}. Influenced by the similar causal statements with the example in Table \ref{tab1} from ATOMIC, the prediction confidence is improved.

\section{Conclusion}
We propose a novel approach, CauSeRL, which could leverage external causal statements to identify the causalities of events. First of all, we design a self-supervised framework to learn context-specific causal patterns from external causal statements. Then, we adopt a contrastive transfer strategy to incorporate the learned context-specific causal patterns into the target ECI model for identification. Experimental results on two benchmarks show that our model achieves the best performance.

\section*{Acknowledgments}
We thank anonymous reviewers for their insightful comments and suggestions. This work is supported by the National Key Research and Development Program of China (No. 2017YFB1002101), the National Natural Science Foundation of China (No.61922085, 61976211). This work is also supported by Beijing Academy of Artificial Intelligence (BAAI2019QN0301) and the joint project with Beijing Baidu Netcom Science Technology Co., Ltd.


\bibliographystyle{acl_natbib}
\bibliography{anthology,acl2021}


\end{document}